\documentclass[conference]{IEEEtran}
\IEEEoverridecommandlockouts
\usepackage{cite}
\usepackage{amsmath,amssymb,amsfonts}
\usepackage{algorithmic}
\usepackage{graphicx}
\usepackage{textcomp}
\usepackage{xcolor}
\usepackage{subfigure}
\def\BibTeX{{\rm B\kern-.05em{\sc i\kern-.025em b}\kern-.08em
    T\kern-.1667em\lower.7ex\hbox{E}\kern-.125emX}}
\begin{document}
\title{Game Level Blending using a Learned Level Representation}
\author{\IEEEauthorblockN{Venkata Sai Revanth Atmakuri}
\IEEEauthorblockA{\textit{Computing Science Department} \\
\textit{University of Alberta}\\
Edmonton, Canada \\
atmakuri@ualberta.ca}
\and
\IEEEauthorblockN{Seth Cooper}
\IEEEauthorblockA{\textit{Khoury College of Computer Sciences} \\
\textit{Northeastern University}\\
Boston, USA \\
se.cooper@northeastern.edu}
\and
\IEEEauthorblockN{Matthew Guzdial}
\IEEEauthorblockA{\textit{Computing Science Department, Amii} \\
\textit{University of Alberta}\\
Edmonton, Canada \\
guzdial@ualberta.ca}
}
\IEEEoverridecommandlockouts
\IEEEpubid{\makebox[\columnwidth]{979-8-3503-2277-4/23/\$31.00~\copyright2023 IEEE \hfill}
\hspace{\columnsep}\makebox[\columnwidth]{ }}
\maketitle
\IEEEpubidadjcol
\maketitle
\begin{abstract}
Game level blending via machine learning, the process of combining features of game levels to create unique and novel game levels using Procedural Content Generation via Machine Learning (PCGML) techniques, has gained increasing popularity in recent years. However, many existing techniques rely on human-annotated level representations, which limits game level blending to a limited number of annotated games. Even with annotated games, researchers often need to author an additional shared representation to make blending possible. In this paper, we present a novel approach to game level blending that employs Clustering-based Tile Embeddings (CTE), a learned level representation technique that can serve as a level representation for unannotated games and a unified level representation across games without the need for human annotation. CTE represents game level tiles as a continuous vector representation, unifying their visual, contextual, and behavioral information. We apply this approach to two classic Nintendo games, Lode Runner and The Legend of Zelda. We run an evaluation comparing the CTE representation to a common, human-annotated representation in the blending task and find that CTE has comparable or better performance without the need for human annotation.
\end{abstract}

\section{Introduction}
Procedural Content Generation via Machine Learning (PCGML) uses machine learning models trained on existing game content to generate various game elements, such as game items, characters, levels, and spaces \cite{summerville2018procedural}. Game level blending is one of the applications of PCGML that has been gaining increasing popularity in recent years \cite{sarkar2020conditional}. It is the process of combining features from different game levels to create novel and unique game levels \cite{guzdial2016learning}. The most popular approach for game level blending in recent years has been using a generative model called a Variational Autoencoder (VAE), which consists of an encoder and decoder network. It attempts to learn a latent space distribution over the training data \cite{kingma2013auto}, encompassing levels from multiple games. By training a VAE on multiple games' levels, we obtain a single, shared latent space distribution, and we can sample from the regions in between the existing games' levels to output blended levels. However, despite advancements in game level blending approaches, most of them still use human-annotated level representations. In addition, these approaches require additional authoring of a unified level representation when dealing with multiple game domains.
If we could find a representation for level blending that didn't require human authoring or annotation we could greatly expand the set of game levels available for blending.

The Video Game Level Corpus (VGLC) has made a significant contribution to PCGML research by providing human-authored level representations for multiple classic game levels \cite{summerville2016vglc}. It offers a text-based tile level representation that assigns each in-game entity in a level to a character or ``tile'', which is then mapped to the actual properties of that element in the game (e.g., ``hazard'' for enemies or other elements that can hurt the player). The VGLC and datasets with similar representations have been the default option for many game level blending projects \cite{sarkar2020conditional}. 
This has been beneficial in terms of allowing for a great deal of research on blending the included game levels. However, it has also been limiting, since researchers can only blend between levels that have been annotated into this representation, which requires human labor. 

Outside of the initial annotation of the dataset, when using representations like the VGLC, authors are required to come up with a unified level representation. This is because the characters represented in one game map to that game's tile properties, and these aren't necessarily shared by a second game. At times, the same character tile is even used to indicate different entities in different games. For instance, the character `B' has the properties `Cannon top', `Cannon', `Solid', and `Hazard' in Super Mario Bros., but `Solid' and `Breakable' in MegaMan, and `Solid' and `Ground' in Lode Runner. Authors therefore must author a mapping that converts all of these tiles to a unified representation. This represents an additional human authoring burden, and one that must be undertaken every time one blends a unique set of games' levels.

There has been a great variety of work exploring game level blending with VAEs \cite{sarkar2020exploring}, including controllable level blending \cite{sarkar2020controllable}, and sequential segment-based level blending \cite{sarkar2020sequential}. While these works have proven effective at demonstrating the uses of VAEs for game level blending tasks, they rely on VGLC-like datasets. Due to the reasons outlined in the above paragraphs, these approaches therefore cannot be extended to new games without significant human effort.

In this paper, we introduce game level blending using Clustering-based Tile Embeddings (CTE). CTE is a learned, unified, domain-independent, and affordance-rich level representation. CTE represent game level tiles in a continuous vector representation, unifying their visual, contextual, and behavioral information. This makes it possible to represent levels from any tile-based game domains \cite{jadhav2022clustering}. We investigate the use of CTE in VAEs for blending game levels of Lode Runner (LR) and The Legend of Zelda (LOZ). We trained a Convolutional Neural Network (CNN) VAE and a Fully Connected (FC) VAE using LR and LOZ game level segments with CTE representation, to compare their performance against baseline PCGML level blending representations. To the best of our knowledge, this is the first approach for game level blending that uses a learned-level representation. 
By investigating the application of CTE to game level blending, we open the possibility of using unannotated games in level blending. Additionally, this allows for blending without additional human authoring to obtain a unified level representation, thereby reducing a significant amount of human labor.

\section{Related Work}

\subsection{Level Blending}

Game level blending is the process of combining different game levels to create levels approximating those from non-existant games. Approaches to game level blending include (1) Snodgrass and Ontañón's domain adaptation, which used training data from some source game to approximate target game training data  \cite{snodgrass2016approach}, (2) Guzdial and Riedl's learned model blending, which employed the original conceptual blending algorithm to combine learned Bayesian networks \cite{guzdial2016learning}, and (3) blending level design constraints from multiple games \cite{cooper2022constraint}.

One popular level blending model in recent years is the Variational Autoencoder (VAE), which can be trained on multiple domains to obtain a single, shared latent space across training data. This means that the regions between different games' levels output blended levels~\cite{sarkar2020controllable,sarkar2020exploring,sarkar2020sequential}. However, these approaches rely on human-annotated level representations and human-authored unified representations.

\subsection{Learned Level Representations}

Learned-level representations are a recent area of PCGML research. This approach involves learning a new representation for game content rather than relying on human annotation or authoring. Approaches in learned-level representation include: (1) entity embeddings, which represent the entities of a game as 25-dimensional continuous vectors. This is achieved by training a Variational Autoencoder (VAE) on a joint representation of raw pixels and dynamic information, similar to tile embeddings~\cite{khameneh2020entity}. Another approach is (2) to train a VQ-VAE on the raw pixels of game levels. A VQ-VAE learns a symbolic-like representation of the game levels by quantizing the input features to a fixed set, which allows users to run level generation approaches that require distinct classes \cite{karth2021neurosymbolic}.

(3) Tile embeddings represent game levels as tiles, with each tile represented as a 256-dimensional continuous vector. This is achieved by training an Autoencoder with a tile's raw pixels and corresponding affordances as input \cite{jadhav2021tile}. Finally, (4) Clustering-based Tile Embeddings (CTE) are an extension to tile embeddings with the addition of edge information and a cluster-based loss to learn a more cohesive latent space~\cite{jadhav2022clustering}.

All of these have proven effective as game level representations. However, only the entity embedding approach has been used in a blending task previously, and that was entity blending not level blending \cite{khameneh2022world}. Thus, none of these learned level representations have been used for game level blending.

For this work, we chose to use a VAE approach with Clustering-based Tile Embeddings (CTE), a learned level representation. This approach could allow game-level blending researchers and developers to explore game-level blending across game domains that do not have an annotated dataset, without relying on a human-authored unified representation.

\begin{figure*}
    \centering
    \includegraphics[width=0.7\textwidth]{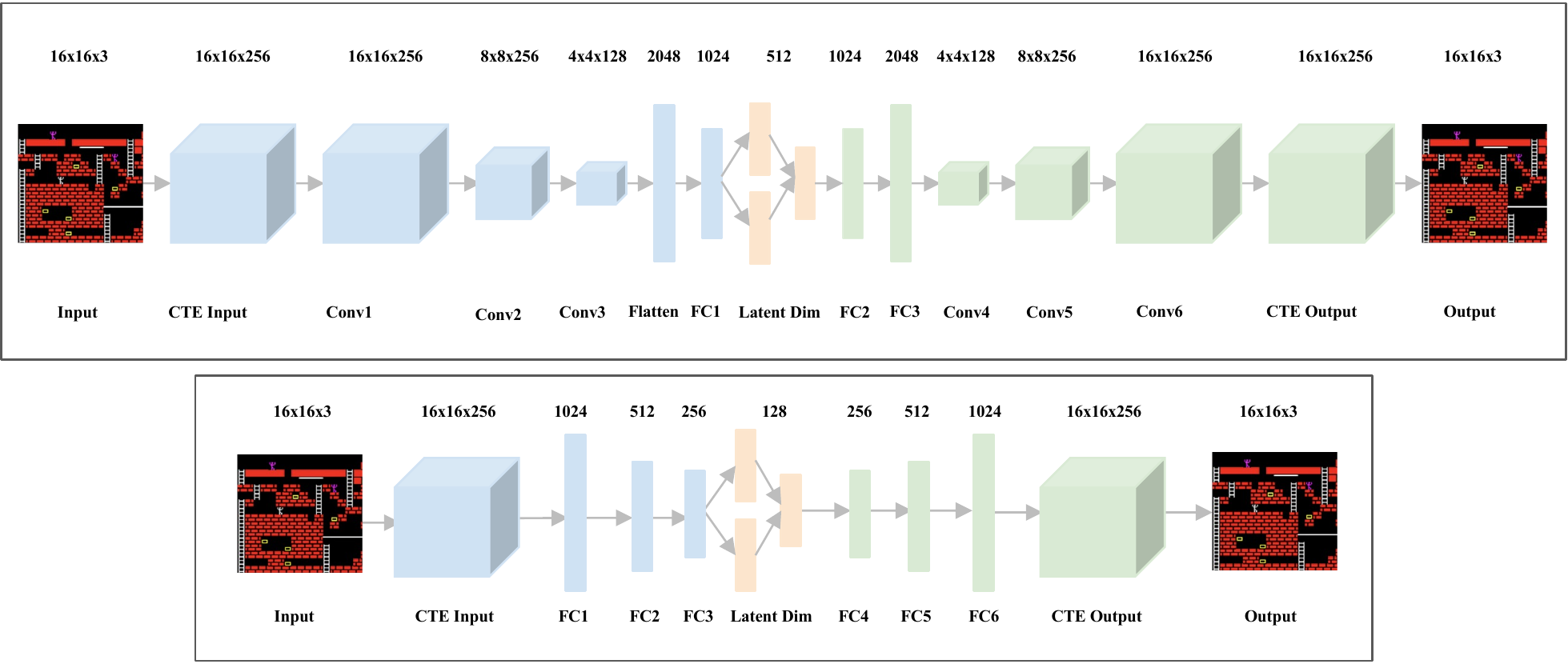}
    \caption{Architecture of CNN and FC Variational Autoencoder models.}
    \label{fig:cnn_architecture}
\end{figure*}

\section{System Overview}

In this paper, we investigate the application of Clustering-based Tile Embedding (CTE), a learned level representation, to game level blending to determine whether we can avoid the labor involved in blending with hand-authored and annotated representations. We use CTE for our level representation and a Variational Autoencoder(VAE) approach for game level blending~\cite{sarkar2020controllable}. For our training set, we chose two classic Nintendo games: The Legend of Zelda (LOZ) and Lode Runner (LR). 
We trained two VAE variants on the CTE level representations for these two games, based on models utilized for level blending with hand-authored representations. Once trained, we used these two models to output blended game level segments.

\subsection{Clustering-based Tile Embedding (CTE)}

We use the Clustering-based Tile Embedding (CTE) level representation because it outperforms tile embeddings \cite{jadhav2021tile} and accounts for affordances, unlike the VQ-VAE approach of Karth et al. \cite{karth2021neurosymbolic}.
CTE is a learned, tile-based level representation that combines the $16\times 16\times 3$ RGB tile pixels with that tile's affordances (e.g., ``Harmful'', ``Solid'', etc.). Each tile embedding is represented by a 256-dimensional vector. This representation can be achieved by training an autoencoder that consists of three inputs: (1) a pixel input of a candidate tile and its neighbors ($48\times 48\times 3$) with the candidate tile ($16\times 16\times 3$) at the center, (2) a 13-dimensional one-hot affordance vector associated with the candidate tile, and (3) $16\times 16$ edge information for the candidate tile \cite{jadhav2022clustering}.  The second input (2) requires manual annotation, but the learned, unified affordances can be used for unannotated game levels \cite{jadhav2022clustering}. A unified level representation can be obtained by training the CTE model with training data from multiple games. We specifically employ a pre-trained CTE supplied by Jadhav and Guzdial \cite{jadhav2022clustering}, as it was trained on a set of games' levels including both LR and LOZ.

\subsection{Dataset}

To create our dataset, we first obtained a unified level representation for The Legend of Zelda (LOZ) and Lode Runner (LR) game levels by using the pre-trained CTE model \cite{jadhav2022clustering}, trained on both game domains. With LR levels represented with dimensions $22\times 32\times 256$, where 22 is the height and 32 is the width of level and 256 is the CTE tile representation, and LOZ levels with dimensions of $h\times w\times 256$ with $h$ is the height and $w$ is the width of level and 256 is the CTE tile representation. 
To train VAE models to achieve level blending using these two game domains, we needed consistent level dimensions, but the level dimensions varied for these two games. To address this problem, we chose the common strategy used in game level blending of splitting the game levels into consistent segments or chunks \cite{sarkar2020controllable,sarkar2020exploring,sarkar2020sequential}. We chose $16\times 16$ as our segment dimensions, representing 16 by 16 tiles, given the similarity to common segment dimensions in the literature and due to our own perceived quality improvements~\cite{sarkar2020controllable,sarkar2020exploring,sarkar2020sequential}. 

For Lode Runner, all 150 game levels available in the VGLC have dimensions of $22\times 32$ tiles. So, we chose to take four $16\times 16$ segments per Lode Runner level, with each segment representing the top-right, top-left, bottom-right, and bottom-left. This meant our data had some overlapping rows between the top and bottom segments, as we felt this would aid generalization. For LOZ, the game levels, or dungeons in this case, vary significantly, but each room across these dungeons has the same dimensions of $11\times 16$ tiles. Therefore, we chose to take each room and apply vertical padding of three blank rows at the top and two blank rows at the bottom to obtain our required $16\times 16$ segment size. Using this representation, we obtained a total of 600 level segments for Lode Runner and 459 level segments for The Legend of Zelda, for a total of 1059 level segments. To train our model, we randomized the total level segments and divided them into three sets: 85\% (900) level segments for training, 10\% (105) level segments for testing, and 5\% (52) level segments for validation. This is a common strategy for training machine learning models \cite{kingma2013auto}.

\subsection{Domains}

We chose The Legend of Zelda (LOZ) and Lode Runner (LR) for this investigation for a number of reasons. First, these two games are included in the VGLC, and therefore PCGML researchers should have some familiarity with them \cite{summerville2016vglc}. Second, these two video games have a roughly equal number of $16\times 16$ level segments, which should benefit a VAE-based blending approach. Third, and most importantly, Lode Runner and the Legend of Zelda represent two very different types of games, with the former being a platformer and the latter a dungeon-crawl/adventure game (as in \cite{sarkar2020conditional}). They also have largely distinct sets of mechanics, only sharing left and right movement. As such, we consider this blending task a ``far transfer'' task or a more difficult blend. With an easier blend (e.g., between two platformer games), we might not be able to get a clear picture of the impact of CTE representation on a level blending task. 

\subsection{Model Architectures}

A Variational Autoencoder (VAE) is a generative model that consists of an encoder and decoder network that attempts to learn a continuous latent space distribution over training data \cite{kingma2013auto}. In this work, we developed two VAE varients: a Convolutional Neural Network VAE (CNN-VAE) \cite{sarkar2020controllable} and a Fully Connected VAE (FC-VAE) \cite{sarkar2020exploring}.

\subsubsection{CNN-VAE}

We visualize our CNN-VAE at the top of Figure \ref{fig:cnn_architecture}.
The CNN-VAE consists of convolutional layers in both the encoder and decoder networks of the VAE. The convolutional layers aim to learn local relationships between multidimensional data, such as images \cite{krizhevsky2017imagenet}, which share similar 2D structural features to levels. For our model, the encoder network consists of three convolutional layers with 256 ($3\times 3$), 256 ($3\times 3$), and 128 ($3\times 3$) filters (filter size), each with a stride value of 2. This is followed by a flatten layer (2048), a fully connected(dense) layer (1024), and a latent sampling layer (512). The decoder network has the same architecture as the encoder network, but in reverse with transpose convolutional layers. Batch normalization and a ReLU activation function are applied after each layer. Batch normalization reduces internal covariance shift, which enables a higher learning rate and fewer training steps. It also acts as a regularizer, preventing overfitting \cite{ioffe2015batch}. This architecture was based around a theoretical understanding of CNN-VAEs given our data size, and then iterated on via visualizing the outputs. 

\subsubsection{FC-VAE}

We visualize our FC-VAE at the bottom of Figure \ref{fig:cnn_architecture}.
We used this second model to better compare the impact of employing a learned level representation against existing game level blending approaches with hand-authored/annotated representations \cite{sarkar2020controllable,sarkar2020exploring,sarkar2020sequential}. The Fully Connected VAE (FC-VAE) consists of fully connected (dense) layers in both the encoder and decoder networks of the VAE. The encoder network consists of 4 fully connected layers with 1024, 512, 256, and 128 dimensions. The decoder network has the same architecture as the encoder network, but in reverse order. Each layer is followed by batch normalization and a ReLU activation function. This architecture was based on an existing VAE model used for level blending \cite{sarkar2020exploring}. Notably, our goal is to not outperform this existing approach, but simply to demonstrate at least equivalent performance with our learned level representation. If we succeed, this will expand the possible game levels available for blending and decrease required human effort.

\begin{figure*}
    \centering
    \includegraphics[width=0.85\textwidth]{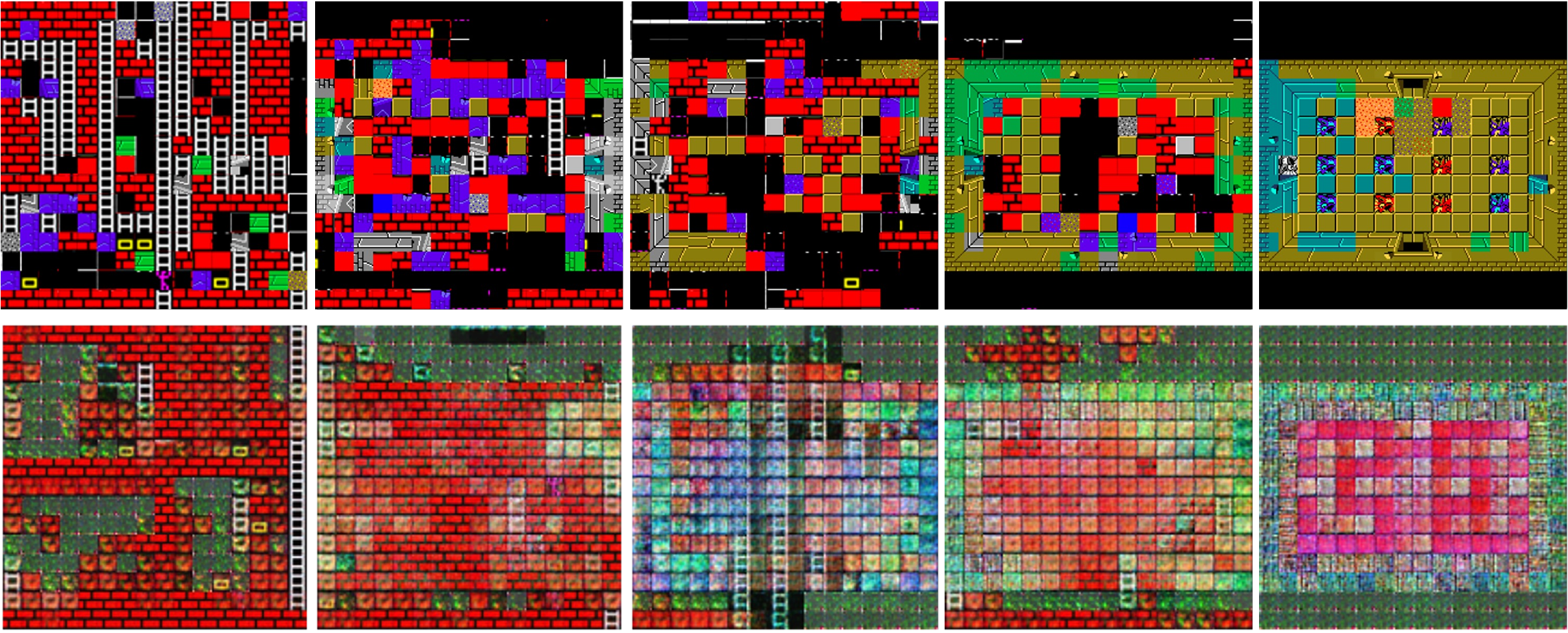}
    \caption{Randomly Blended Segments Visualization: (Top) Nearest Actual Tile Embedded (Top) and (Bottom) CTE Decoder.}
    \label{fig:visualizations}
\end{figure*}

\subsection{Training}
All models presented in this paper, including the baselines, were trained using the Adam optimizer \cite{kingma2014adam} with a learning rate of 1e-5, following the training procedure outlined below.

VAE loss consists of two loss terms: the reconstruction loss and the variational loss (also known as the KL loss). Although our training set represents a fairly equal number of segments, with 600 segments representing Lode Runner and 459 segments representing the Legend of Zelda, the Legend of Zelda segments had a more consistent segment representation compared to the Lode Runner segments, which have higher structural variance. This caused an imbalanced dataset issue. To address this problem, we employed a weighted loss \cite{wang2016training} for our training, by weighting the reconstruction loss of Lode Runner segments with 0.57 (600/1059) and the Legend of Zelda segments with 0.43 (459/1059). In addition we multiplied the reconstruction loss by a constant value of 4, which we found improved the reconstruction accuracy during training.

For the variational loss, to overcome the issue of KL vanishing, we used cyclic annealing. We started with a value of 0 for the first cycle of 200 epochs to avoid the KL vanishing issue. Then, we used cyclic annealing for the variational loss weight rising from 0 to 0.01 for the first half cycle of 100 epochs, increasing by 0.0001 for each epoch. We then kept the variational loss weight at a constant value of 0.01 for the next half cycle of 100 epochs. This procedure continued for each cycle of 200 epochs \cite{fu2019cyclical}.

We trained the models for a total of 4000 epochs. Although we could have employed a higher learning rate value due to the use of batch normalization in our models, we chose to use a learning rate of 1e-5 to achieve better generalization performance and to avoid local minima \cite{smith2017cyclical}. This is consistent with other PCGML approaches that train for long periods with low learning rates to deal with high variance level data \cite{summerville2017understanding}. 

\subsection{Model Usage}

\subsubsection{Generation and Blending}
After training a model, we can generate randomly blended or unblended level segments or blend existing level segments. To randomly generate blended or unblended novel level segments, we sample from the latent space. We pass this latent space sample to the decoder to obtain a novel level segment in the CTE representation \cite{kingma2013auto}. Since this is a random sample, we cannot guarantee if the output level segment will be a mix of the games' levels (blended) or not (unblended).

The procedure for blending existing level segments is similar to randomly blending game level segments. Instead of random input, we pass existing level segments (contained in the training data) that we want to blend to the trained encoder. This allows us to obtain their locations in the learned latent space. We can then sample from between these points to identify various blends between the existing level segments. This process is referred to as interpolation \cite{dumoulin2016adversarially}. We can identify various points along this interpolation and pass them to the trained decoder to obtain a novel level segment that represents a blend between pairs of inputs.

\subsubsection{Visualizing Output}
In our approach, we use CTE, a learned representation, as our level representation. CTE represents each $16\times 16$ pixel image as a tile, and each tile is represented as a 256-dimensional continuous vector. Once we train our VAE model on the CTE level representation, the output of our VAE model is also represented in the CTE representation. To obtain a graphical representation of the level segment, we can use two approaches: (1) nearest actual tile embedding or (2) CTE Decoder visualization. We visualize both approaches in Figure \ref{fig:visualizations}. In this paper, we primarily use the nearest actual tile embedding approach to evaluate our game blending approach.

The nearest actual tile embedding approach involves relying on the positions of the original training data in the learned latent space of the trained CTE model. By feeding the original inputs into the CTE model after training, we can extract a mapping of embeddings to $16\times 16$ pixels. After outputting a blend, we can then employ the Annoy Library to find the closest actual tile embedding and output the associated $16\times 16$ pixel representation. For the nearest actual tile embedding approach we employ the Manhattan distance to find the closest embedding from the training dataset. This has been the most common approach for visualizing the output of level generators trained with tile embeddings \cite{jadhav2021tile}.

Visualizing the output with the CTE Decoder involves obtaining the trained VAE model output in the CTE representation, with each tile represented by a 256-dimensional continuous vector. By using the trained CTE decoder we can convert each of these 256-dimensional values to the CTE decoder output: $16\times16$ pixels, a 13-length vector representing the affordances, and the edge information. This approach allows us not only to obtain pixels for each tile, but also to output novel tile visualizations and novel sets of affordances. Therefore this approach can allow us to move beyond the original tiles from the two games and to output novel tiles. To the best of our knowledge, this represents the first time novel tiles have been generated as part of a level blending approach. However, because of their novelty, its not possible to evaluate them against exiting tiles. In addition, as we demonstrate in Figure \ref{fig:visualizations}, these novel tiles do not match our human expectations in terms of pixel art quality at this time.

\section{Evaluation}

The main aim of this paper is to investigate the use of learned level representations in game level blending, as opposed to traditional human-annotated level representations. We cannot employ the CTE Decoder visualization to evaluate our approach since it outputs novel tiles, making comparison impossible without a human evaluation or similar approach. Given that we want to run a large-scale evaluation with multiple comparisons between our learned representation and a human-authored level representation, a human subject study is infeasible. Therefore, for our evaluation, we employ the nearest actual tile embedding visualization and a series of metrics for the purpose of comparison.

For our evaluation, we used two common approaches \cite{jadhav2022clustering,sarkar2020sequential,sarkar2020controllable} for PCGML level generation research: (1) tile-based metrics typically employed in expressive range analysis \cite{summerville2017understanding}: density, non-linearity, leniency, interestingness, and path proportions, and (2) a playability analysis, using A* agents to test the playability of the generated and blended game segments.

\subsection{Baselines}

For our baseline evaluation, we created a hand-authored unified level representation to represent the VGLC levels of both Lode Runner and Legend of Zelda games. This unified level representation was obtained using techniques similar to those used in previous game level blending approaches \cite{sarkar2020controllable,sarkar2020sequential}. 
We combined the repeated tile characters in both games and represented them as a single character that combined both games' tile affordances, as shown in Table \ref{tab:baselineunifiedlevelrepresentation}.
\begin{table}[htbp]
\begin{center}
\begin{tabular}{|l|c|}
\hline
\textbf{Tile Properties} & \textbf{Text} \\
\hline
Solid, Ground, Block & `B' \\
Solid, Diggable, Ground & `b' \\
Passable, Empty & `.' \\
Passable, Climable, Rope, Empty & `-' \\
Passable, Climable, ladder & `\#' \\
Passable, Pickupable, Gold & `G' \\
Damaging, Enemy & `E' \\
Damaging, Spawn, Solid, Hazard & `M' \\
Solid & `F' \\
Element & `P' \\
Element, Block & `I' \\
Element, Solid & `O' \\
Solid, Openable & `D' \\
Passable, Climbable & `S' \\
Solid, Wall & `W' \\
\hline
\end{tabular}
\end{center}
\caption{Baseline Unified Tile Representation.}
\label{tab:baselineunifiedlevelrepresentation}

\end{table}
Hand-authoring this unified tile representation represents the standard approach in PCGML level blending research. To compare it to our choice of the CTE representation, we trained the same two VAE models presented in the system overview section, with the same training procedure. The only difference was in the reconstruction loss, where we employed Mean Square Error (MSE) for the CTE representation, but Cross-entropy Error for the baseline representation given baseline's use of a one-hot encoding of tiles. With this setup, if our CTE approach has even equivalent performance to the hand-authored unified tile representation then we can safely argue that we can employ CTE for PCGML level blending. 
If so, this means we can avoid the requirement of human authoring or annotation to obtain a unified level representation for game level blending, and more unannotated games can be used in game level blending.

\subsection{Tile-based Metrics}
Tile-based metrics are the most commonly utilized approach to evaluate generated randomly blended and blended segments~\cite{jadhav2022clustering,sarkar2020sequential}. Originally developed for expressive range analysis \cite{smith2010analyzing}, they capture various properties of a distribution of levels and provide insights to developers and researchers. 
In this work, we use density, non-linearity, leniency, interestingness, and path-proportion as our tile-based metrics, which we adapt from prior work and describe in more detail below. In all cases we drew on existing implementations of these metrics.

\subsubsection{Density}
Measures the proportion of segments that are occupied by tiles that the player can stand on, such as solid ground(\includegraphics[width=0.3cm]{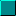}, \includegraphics[width=0.3cm]{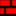}, \includegraphics[width=0.3cm]{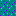}), blocks (\includegraphics[width=0.3cm]{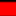}), etc. \cite{sarkar2020sequential}.

\subsubsection{Non-Linearity}
Measures how well a segment fits a line. It is calculated as the mean square error (MSE) measured by fitting a line to the topmost non-empty tile of columns in a segment using linear regression\cite{sarkar2020exploring, smith2010analyzing}

\subsubsection{Leniency}
Measures the proportion of segments that are not occupied by tiles that have enemy \includegraphics[width=0.3cm]{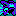} or hazard \includegraphics[width=0.3cm]{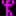} properties. This is a rough approximation of difficulty \cite{sarkar2020sequential}.

\subsubsection{Interestingness}
Measures the proportion of segments that are occupied by tiles that are interesting or otherwise unusual \cite{summerville2017understanding}. For Lode Runner and Legend of Zelda, the interesting tiles are gold \protect\includegraphics[width=0.3cm]{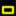}, climbable (\includegraphics[width=0.3cm]{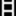}, \includegraphics[width=0.3cm]{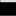}, \includegraphics[width=0.3cm]{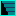}), door \includegraphics[width=0.3cm]{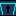}, and elements \includegraphics[width=0.3cm]{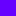}.

\subsubsection{Path-Proportion}
Measures the proportion of segments that are occupied by tiles the player can stand on or otherwise occupy \cite{sarkar2020sequential}. In comparison to playability, which we also measure, this allows us to approximate the proportion of tiles the player can reach.

\subsection{Playability Metrics}

Playability is a metric that evaluates whether a player can move through a level segment. One of the core requirements of game level generation generally is to achieve playability, and A* pathfinding agents are often used to approximate it. However, blended and random blended levels have tiles from both games, each representing their own game affordances. This complicates the playability evaluation problem. Therefore, we developed two A* agents, an A* agent configured for each game: Lode Runner (LR) and The Legend of Zelda (LOZ). They are described in more detail below.

\subsubsection{Lode Runner (LR) A* Agent} The LR A* agent is configured to take all the possible actions a real-world player can take in LR, with gravity taken into consideration. The actions include (1) Moving left or right when there are ground tiles and the neighboring tiles are empty and passable. (2) Falling down from an action of moving left or right when the neighboring tile is empty or passable but there is no ground tiles below. (3) Digging through the breakable tiles, and (4) Climbing the ladder to move up or down and using the rope to move left or right.

\subsubsection{Legend of Zelda (LOZ) A* Agent} Our LOZ A* agent is configured to take all the possible actions that a real-world player can take in LOZ. By default, we assume the player doesn't have a ladder to cross the element tiles \includegraphics[width=0.3cm]{P.png}.

\subsubsection{Playability Evaluation}

For our playability evaluations, as we mentioned earlier, blended level segments have tiles from both games, each representing their own game affordances. Therefore, we measured the playability of the given set of segments for both the LR A* agent and LOZ A* agent configured as outlined above.

For the LR A* agent playability evaluation, a given level segment might represent either a more LR-like segment or a more LOZ-like segment. To address this problem, we categorized the set of level segments into two categories: either LR-like segments or LOZ-like segments, based on the tiles represented in the segments. A segment with a majority of the tiles representing LR tiles is categorized as a LR-like segment, and a segment with a majority of the tiles representing LOZ tiles is categorized as a LOZ-like segment.

We then ran our configured LR A* agent on all the LR-like level segments with the goal of collecting the gold presented in the level segment, as it's the main way we can evaluate the playability of a LR level segment, since that's the game's goal. We gave two starting positions towards this goal because we found during our evaluation that the LR A* agent could get stuck between blocks on some of the LR-like segments if we only had one starting position from the top of the level segment. Therefore, we also gave a second starting position from the bottom and found that this solved the issue. We consider an LR-like segment to be playable by an LR A* agent if the agent can collect at least half of the gold tiles presented in the level segments from either starting position. These simplifications were included since each segment is only a subsection of an LR level, and we could therefore imagine other paths ``offscreen''.

We also ran our configured LR A* agent on all the LOZ-like level segments. For these, we gave four start and goal positions. The four starting positions and corresponding goal positions were: (1) starting from the bottom-left of the room segment with a goal to reach the top-right, (2) starting from the bottom-right with a goal to reach the top-left, (3) starting from the top-left with a goal to reach the bottom-right, and (4) starting from the top-right with a goal to reach the bottom-left. We consider a segment to be playable if there is a path between at least two of the above four goals because, if any two of the above goals are satisfied, this indicates that the agent can move through the room, which satisfies the player goals for LOZ. We didn't just use the door sprites as our start and goal locations as there was no guarantee a LOZ-like segment would include doors at particular positions.

\begin{table*}[htbp]
\begin{center}
\begin{tabular}{|l|c|c||c|c|}
\hline
\textbf{Metrics} & \textbf{CTE FC-VAE} & \textbf{VGLC FC-VAE} & \textbf{CTE CNN-VAE} & \textbf{VGLC CNN-VAE} \\
\hline
Density & $0.3183\pm0.1227$ & $0.3544\pm0.1131$ & $0.3991\pm0.1227$ & $0.3304\pm0.0479$ \\
\hline
Non-linearity & $0.1551\pm 0.2152$ & $0.1686\pm0.2235$ & $0.1639\pm0.2339$ & $0.0667\pm0.1567$ \\
\hline
Leniency & $0.9907\pm 0.0253$ & $0.9874\pm0.0358$ & $0.9864\pm0.0269$ & $0.9951\pm0.0186$ \\
\hline
Interestingness & $0.0992\pm0.0788$ & $0.0949\pm0.0692$ & $0.0919\pm0.0717$ & $0.0292\pm0.0338$ \\
\hline
Path Proportion & $0.6416\pm0.1311$ & $0.6467\pm0.1120$ & $0.6496\pm0.1177$ & $0.6545\pm0.0574$ \\
\hline
E-Distance & \textbf{0.04016} & 0.1755 & 0.07036 & 0.1930 \\
\hline
LR A* & \textbf{33.5} & 32.0 & 32.0 & 31.5 \\
\hline
LOZ A* & 79.0 & 79.5 & \textbf{81.5} & 79 \\
\hline
\end{tabular}
\end{center}
\caption{Tile-Based metrics results are presented with $mean\pm standard deviation$, as well as E-Distance and Playability results for 1000 randomly blended novel level Segments for each Approach}
\label{tab:generatedvstrainedresults}
\end{table*}

For the LOZ A* agent playability evaluation, we didn't need to categorize the set of segments into two categories because the LOZ A* agent doesn't have a gravitational constraint and can move freely through the level segments. Therefore, we evaluated it with the same four start and goal positions as in the LR A* agent playability evaluation for LOZ-like segments but with the LOZ A* agent.

\section{Results}

Table \ref{tab:generatedvstrainedresults} presents the results for 1000 randomly blended novel level segments for each of the approaches we presented in this paper, labeled as `CTE FC-VAE': our Fully connected VAE trained using the CTE representation, `VGLC FC-VAE': the Fully connected VAE trained using VGLC representation, `CTE CNN-VAE': our CNN-VAE trained using the CTE representation, and `VGLC CNN-VAE': the CNN-VAE trained using the VGLC representation. It contains three different groups of evaluation metrics: (1) the Tile-based results presented in the evaluation section, represented as mean $\pm$ standard deviation values over the 1000 randomly blended segments. (2) the E-distance between the 1000 randomly blended segments and the training segments for the presented metrics~\cite{summerville2017understanding}. (3) the Playability results of the 1000 segments for each A* agent, labeled `LR A*': for the Lode Runner A* agent, and `LOZ A*': for The Legend of Zelda A*.

For the tile-based results presented in Table \ref{tab:generatedvstrainedresults}, all the results are similar for CTE FC-VAE and VGLC FC-VAE.
However, the results vary slightly for our CTE CNN-VAE and the VGLC CNN-VAE, with the main difference being in interestingness, non-linearity and density. The higher mean range indicates that the CTE CNN-VAE has more tiles representing each metric compared to the VGLC CNN-VAE.
All of the remaining results were similar, suggesting that both VAE variants have similar performance in terms of tile-based metrics between the CTE and VGLC representations. This implies the CTE representation could be used instead of human annotated level representations without negatively impacting blending results.

E-distance is a measure of similarity between different distributions \cite{szekely2013energy} and has been used as an evaluation metric for generative models \cite{summerville2018expanding}. A lower E-distance value indicates that the measured distributions are more similar. In this work, E-distance is measured for the 1000 randomly blended segments for each approach against the training segments, and the results are bolded for the best values in Table \ref{tab:generatedvstrainedresults}. The E-distance results vary between the CTE and VGLC representations for the two variants. The CTE FC-VAE and CTE CNN-VAE 
both have lower E-distance values compared to the VGLC FC-VAE and VGLC CNN-VAE, suggesting that the models trained on CTE have outputs more similar to the training segments.

Finally, in general, the playability results are similar for all the approaches. The playability is better in all cases for the LOZ A* agent, since it doesn't have gravity constraints. Whereas for the LR A* agent, the playability score is approximately 32\%, since it has more constraints. That said, the CTE approaches have a slightly better playability than the VGLC approaches, which further supports the E-distance results.

\begin{figure*}
    \centering
    \includegraphics[width=0.95\textwidth]{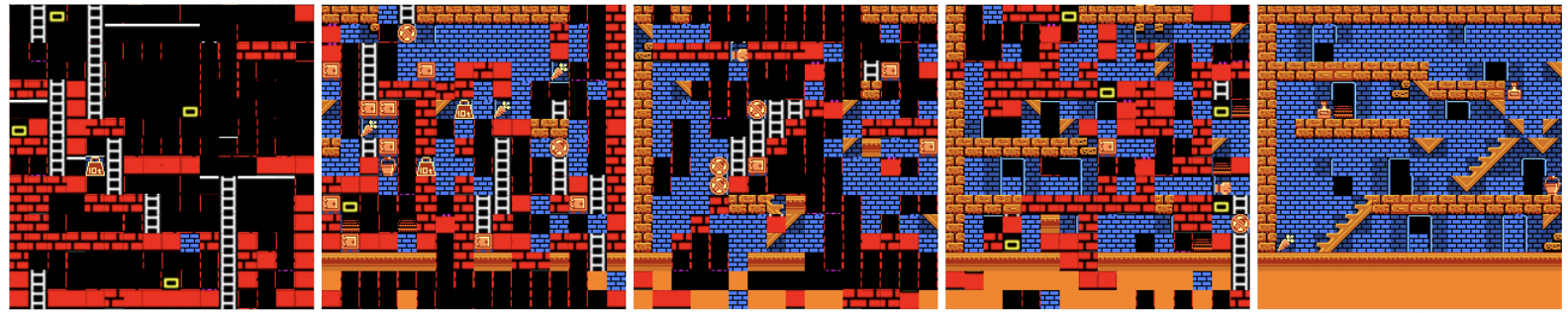}
    \caption{Randomly Blended novel level segments between Bugs Bunny Crazy Castle(game without annotated dataset) and Lode Runner using CTE.}
    \label{fig:RandomBlend}
\end{figure*}

\section{Discussion and Future Work}

Our results from Table \ref{tab:generatedvstrainedresults} support that our approach, which uses CTE and doesn't require human annotation or authoring, is largely similar and, in some cases, better (as measured by E-distance and playability) than using VGLC, which did require human annotation or authoring. In addition to the results from Table \ref{tab:generatedvstrainedresults}, we have also included tile-based metric results for level blending with different percentages of each level segment type as line graph in our supplementary material. These graphs further support the results from Table \ref{tab:generatedvstrainedresults} for level blending, showing a similar tread for each metric.

Our initial motivation for this work was to use unannotated game domains with the CTE representation for game blending. However, the CTE representation has never been used for game blending before, and therefore has never been evaluated. In addition, we are unable to evaluate unannotated game domains in the same manner as the above as there is no ground truth available. Therefore, for this work, we chose to use CTE representation for game domains that had annotated datasets available. But as we outlined in our introduction, using CTE would expand the available game domains since it can be used with unannotated game domains, which is one of our long-term goals. As proof of concept, we also included random blend level segments in Figure \ref{fig:RandomBlend}. These were trained on Bugs Bunny Crazy Castle (an unannotated game) and Lode Runner using the CNN-VAE model with the CTE representation, following the same training procedure outlined in our training section. The results included in Figure \ref{fig:RandomBlend} further show that the model is capable of producing random blended segments that are similar to Lode Runner (the left segment of the figure) and similar to Bugs Bunny Crazy Castle (the right segment of the figure), as well as the in-between blended segments (the middle segments of the figure). This demonstrates the possibility of game blending for an unannotated dataset using the CTE representation.

There are many avenues we hope to explore in future work. One clear vector is evaluating the use of the CTE decoder as the visualization approach, since it provides blended game levels with a blended tile representation that includes blended affordances as shown in Figure \ref{fig:visualizations}. This could potentially lead to novel game levels that implicitly encode novel gameplay mechanics, through extending the approaches developed by Summerville et al. \cite{summerville2020extracting}. Another possible area would be to evaluate our approach for game domains without annotated datasets. This could allow us to approximate target game levels for games without any level training data available, such as in-development games.  Another possible area would be further exploring the possibility of CTE by increasing the number of game domains beyond just two game domains, since our approach doesn't require human annotation/authoring to achieve a uniform/unified level representation. Although we used a VAE as our game blending approach in this work, further experiments need to be conducted to further test CTE with other game blending approaches in order to determine what models best support CTE-based level blending.

\section{Conclusions}

In this paper, we introduce a novel approach to game level blending that uses Clustering-based Tile Embeddings (CTE). CTE is a learned level representation that can be used for unannotated game domains and does not require human annotation or authoring to achieve a unified level representation for game level blending. We provide an evaluation of this approach against a baseline game level blending representation and show that it performs equivalently to or better than this baseline without the need for human annotation. We also provide examples of blended levels that include novel, blended tiles and blended levels between annotated and unannotated games. This work has greatly expanded the domains available for level blending while decreasing required human labor.

\bibliography{conference_101719}

\end{document}